\documentclass[pdflatex,sn-mathphys-num]{sn-jnl}

\usepackage{graphicx}%
\usepackage{multirow}%
\usepackage{amsmath,amssymb,amsfonts}%
\usepackage{amsthm}%
\usepackage{mathrsfs}%
\usepackage[title]{appendix}%
\usepackage{xcolor}%
\usepackage{textcomp}%
\usepackage{manyfoot}%
\usepackage{booktabs}%
\usepackage{algorithm}%
\usepackage{algorithmicx}%
\usepackage{algpseudocode}%
\usepackage{listings}%
\usepackage{amssymb}
\usepackage{multirow}
\usepackage{booktabs}
\usepackage{float}
\usepackage{etoolbox}
\newcommand{\etal}{\textit{et al.}}

\theoremstyle{thmstyleone}%
\theoremstyle{thmstyletwo}%

\theoremstyle{thmstylethree}%

\begin{document}

\title[Article Title]{Multi-scale Attention-Guided Intrinsic Decomposition and Rendering Pass Prediction for Facial Images}


\author*{\fnm{Hossein} \sur{Javidnia}}\email{hossien.javidnia@tcd.ie}

\affil{\orgdiv{Electronic and Electrical Engineering}, \orgname{Trinity College Dublin}, \orgaddress{\city{Dublin}, \country{Ireland}}}


\abstract{Accurate intrinsic decomposition of face images under unconstrained lighting is a prerequisite for photorealistic relighting, high-fidelity digital doubles, and augmented-reality effects. This paper introduces MAGINet, a Multi-scale Attention-Guided Intrinsics Network that predicts a $512\times512$ light-normalized diffuse albedo map from a single RGB portrait. MAGINet employs hierarchical residual encoding, spatial-and-channel attention in a bottleneck, and adaptive multi-scale feature fusion in the decoder, yielding sharper albedo boundaries and stronger lighting invariance than prior U-Net~\cite{ronneberger2015u} variants. The initial albedo prediction is upsampled to $1024\times1024$ and refined by a lightweight three-layer CNN (RefinementNet). Conditioned on this refined albedo, a Pix2PixHD-based~\cite{wang2018Pix2PixHD} translator then predicts a comprehensive set of five additional physically based rendering passes: ambient occlusion, surface normal, specular reflectance, translucency, and raw diffuse colour (with residual lighting). Together with the refined albedo, these six passes form the complete intrinsic decomposition. Trained with a combination of masked-MSE, VGG, edge, and patch-LPIPS losses on the FFHQ-UV-Intrinsics dataset~\cite{dib2024mosar}, the full pipeline achieves state-of-the-art performance for diffuse albedo estimation and demonstrates significantly improved fidelity for the complete rendering stack compared to prior methods. The resulting passes enable high-quality relighting and material editing of real faces.}




\maketitle

\section{Introduction}\label{sec:intro}
\noindent
The creation of photorealistic, editable digital humans is a foundational challenge in computer vision and graphics, with transformative potential for filmmaking, virtual reality, and next-generation communication platforms. A critical bottleneck in democratizing this technology is the ability to perform inverse rendering: recovering the intrinsic properties of a human face---its 3D shape, surface reflectance, and the effects of illumination---from a single, unconstrained portrait captured with an ordinary camera. Without specialized hardware such as light stages~\cite{debevec2000acquiring}, this task is profoundly ill-posed, requiring computational models to disentangle a multitude of interacting physical phenomena from a single 2D observation.

\noindent
To tackle this challenge, the field has recently converged on two powerful paradigms. The first involves implicit volumetric representations, where methods based on 3D Generative Adversarial Networks (GANs) or Neural Radiance Fields (NeRFs) learn to synthesize a continuous, 3D-aware model of the face from 2D image collections~\cite{mei2024holo}. These approaches can produce stunningly realistic, view-consistent avatars that can be relit. The second paradigm is driven by generative relighting models, which leverage the power of diffusion models to re-render a portrait under novel lighting conditions. Often trained in a self-supervised manner, these models excel at producing photorealistic results and handling complex effects like cast shadows without requiring explicit 3D geometry or paired ground-truth data~\cite{ponglertnapakorn2023difareli}.

\noindent
While these implicit and generative approaches achieve state-of-the-art photorealism, they present critical trade-offs for content creation workflows. Volumetric models~\cite{mei2024holo} can be computationally intensive, and their implicit nature means they do not natively produce the standard, editable rendering passes---such as separate specular, normal, or ambient occlusion (AO) maps---that are the lingua franca of professional graphics pipelines. Similarly, diffusion-based relighters~\cite{ponglertnapakorn2023difareli} typically operate as holistic image-to-image translators, producing a final, ``baked'' relit image without the intermediate, interpretable assets required for flexible material editing or compositing. This reveals a clear and unmet need in the field: a practical method that can decompose a single portrait into a complete and explicit stack of Physically Based Rendering (PBR)-compatible passes, bridging the gap between pure synthesis and the practical demands of downstream editing applications.

\noindent
Our key insight is that this highly ill-posed, multi-component decomposition problem can be made tractable by structuring it as a progressive, multi-stage pipeline that strategically disentangles illumination removal from the data-driven synthesis of geometric and material properties. We introduce MAGINet, a three-stage framework designed for this purpose. First, a \textbf{M}ulti-scale \textbf{A}ttention-\textbf{G}uided \textbf{I}ntrinsics \textbf{Net}work---the core MAGINet---leverages a deep receptive field and a dual-attention bottleneck to predict a robust, light-normalized diffuse albedo map. Second, a lightweight RefinementNet upsamples and sharpens this albedo. Finally, conditioned on this clean albedo, a conditional GAN-based translator~\cite{wang2018Pix2PixHD} synthesizes the five remaining PBR passes: surface normals, specular reflectance, ambient occlusion, translucency, and raw diffuse color.

\newpage
\noindent
\textbf{Contributions.} Our work makes the following contributions:
\begin{enumerate}
    \item We propose a multi-stage pipeline that, to our knowledge, is the first single-image method designed to explicitly predict a complete six-pass PBR stack (albedo, normals, AO, specular, translucency, and raw diffuse color), providing assets suitable for direct use in standard rendering workflows.
    \item We introduce a specialized multi-scale attention network for albedo estimation and demonstrate that a conditional GAN framework can successfully learn to generate geometrically-plausible passes (e.g., normals, AO) from a purely 2D albedo map, exploiting the strong statistical correlations present in facial data.
    \item Our full pipeline achieves state-of-the-art performance for diffuse albedo estimation on the FFHQ-UV-Intrinsics benchmark~\cite{dib2024mosar} and is validated on a synthetic dataset with absolute ground-truth geometry, where it achieves a mean angular error of just $2.8^{\circ}$ for surface normals, confirming the physical plausibility of our decomposed passes.
\end{enumerate}
\vspace{0.4cm}

\section{Related Work}\label{sec:lit}
\noindent
Our work sits at the intersection of classical intrinsic image decomposition, modern generative models for facial relighting, and physically-based inverse rendering. We position our contribution by reviewing the evolution and trade-offs within each of these domains.

\subsection{Intrinsic Decomposition: From Classical Priors to Deep Learning}
\noindent
Intrinsic image decomposition (IID) is the classic, ill-posed problem of separating an image into its constituent layers, primarily a view-invariant reflectance (albedo) and a view-dependent shading map~\cite{barrow1978recovering}. Early methods relied on hand-crafted priors. With the advent of deep learning, data-driven approaches became dominant. Methods like SfSNet~\cite{sengupta2018sfsnet} and InverseFaceNet~\cite{kim2018inversefacenet} pioneered the use of convolutional neural networks (CNNs), often trained on synthetic data, to predict shape, albedo, and simple spherical harmonics (SH) lighting from a single image. These foundational works demonstrated that deep networks could learn powerful priors for disentanglement. However, these early approaches were often constrained by simplifying assumptions, such as Lambertian surface reflectance and low-frequency lighting models. This limited their ability to faithfully represent real-world faces, which exhibit complex, non-Lambertian effects. Consequently, their outputs---typically just a diffuse albedo and a coarse shading map---lack the necessary components for high-fidelity, physically-based relighting. Our work moves beyond this limited two-component decomposition, aiming to recover a far richer set of six PBR passes.

\subsection{Generative Models for Implicit Face Relighting}
\noindent
A more recent and powerful trend in the literature bypasses explicit decomposition altogether, instead framing relighting as an image synthesis task solved by generative models. This paradigm can be broadly divided into 2D image-space methods and 3D-aware volumetric methods.

\noindent
\textbf{2D Generative Relighting.} Diffusion models have emerged as a state-of-the-art tool for photorealistic image synthesis. DiFaReli~\cite{ponglertnapakorn2023difareli} leverages a conditional diffusion model to relight faces in the wild by manipulating a disentangled lighting code. A key advantage of this approach is its ability to train in a self-supervised fashion on 2D image collections, obviating the need for expensive light stage data or 3D ground truth. However, the output of such models is a final, baked relit image. They do not produce the intermediate, editable PBR assets that are essential for artistic control and integration into standard graphics pipelines.

\noindent
\textbf{3D-Aware Volumetric Relighting.} Concurrently, advances in 3D-aware GANs and NeRFs have enabled the creation of relightable 3D facial avatars from single or multiple images. Holo-Relighting~\cite{mei2024holo}, for example, inverts a single image into the latent space of a pretrained 3D GAN and then uses a specialized module to render the resulting volumetric representation under novel lighting. These methods achieve impressive 3D consistency and can model complex, view-dependent effects. The primary trade-off is that the scene representation is implicit---contained within the network weights and a latent code---and does not naturally afford access to explicit, layered PBR maps.

\noindent
In contrast to these implicit synthesis approaches, our work pursues an explicit decomposition strategy. While implicit methods may achieve superior photorealism for a single target lighting condition, our approach provides an intermediate, interpretable, and editable set of PBR assets, offering greater flexibility for artists and downstream applications.

\subsection{Towards Full PBR and Physically-Based Inverse Rendering}
\noindent
Our goal of predicting a full PBR stack aligns with the broader ambition of physically-based inverse rendering. This field seeks to recover geometry, materials, and lighting from images to enable applications like realistic object insertion and material editing~\cite{pandey2021total, lin2025iris}. Works like Total Relighting~\cite{pandey2021total} have developed complete systems for portrait relighting, often using deep learning to predict per-pixel lighting representations. More recent efforts in inverse rendering for general scenes, such as IRIS~\cite{lin2025iris}, can recover HDR lighting and physically-based materials but typically require multi-view LDR or HDR inputs, making them ill-suited for our target scenario of a single, unconstrained LDR portrait. Our work occupies a specific and challenging niche: we adopt the ambitious goal of full PBR pass prediction from the inverse rendering community but tailor our approach to the highly constrained, single-image facial domain.

\subsection{The Role of Supervision in Facial Inverse Rendering}
\noindent
The choice of supervisory signal represents a fundamental design axis in this domain. Light stage capture provides the highest-fidelity ground truth, recording a subject under hundreds of controlled lighting conditions~\cite{debevec2000acquiring, mei2024holo}. However, this is expensive and lab-constrained. At the other end of the spectrum, self-supervised methods offer great flexibility by training on in-the-wild data~\cite{ponglertnapakorn2023difareli}, but controlling the physical plausibility of the disentangled components can be challenging. We opt for a pragmatic middle ground by training on the FFHQ-UV-Intrinsics dataset~\cite{dib2024mosar}. We acknowledge that this data constitutes pseudo-ground-truth, as it is itself the output of an inverse-rendering pipeline and thus inherits any biases from that system. However, it provides a crucial, strong supervisory signal necessary for learning the complex, non-linear correlations between a 2D albedo map and geometric properties like normals and ambient occlusion---a task that would be extremely difficult in a purely self-supervised setting.

\section{Methodology}\label{sec:method}

\subsection{Problem Setting}

Let \(I\!\in\!\mathbb{R}^{3\times1024\times1024}\) denote an aligned, front‐facing portrait captured under unknown illumination.
The goal is twofold:  
(i) recover a light–normalised diffuse albedo \(\mathbf{A}\) that is free from shading and specular interference, and  
(ii) predict a set of physically–based rendering passes  
\(\mathbf{R}= \{\mathbf{R}^{\text{AO}}, \mathbf{R}^{\text{N}}, \mathbf{R}^{\text{S}}, \mathbf{R}^{\text{T}}, \mathbf{R}^{\text{D}}\}\),  
where superscripts denote ambient occlusion, surface normal, specular, translucency, and raw diffuse colour (with residual lighting), respectively.  
Reliable estimation of these layers enables downstream relighting, material editing, and neural rendering.

\vspace{2pt}
\subsection{Three-stage Decomposition Pipeline}

Our decomposition pipeline consists of three stages, processing information across distinct scales. First, MAGINet predicts an initial $512\times512$ albedo by jointly leveraging global context and local texture. This $512\times512$ output is then upsampled to $1024\times1024$ resolution before being passed to a RefinementNet, which sharpens and refines the prediction. Finally, a Pix2PixHD~\cite{wang2018Pix2PixHD} network translates the high-resolution refined albedo into the five target rendering passes.

\begin{figure}
\centering
\includegraphics[scale=0.2]{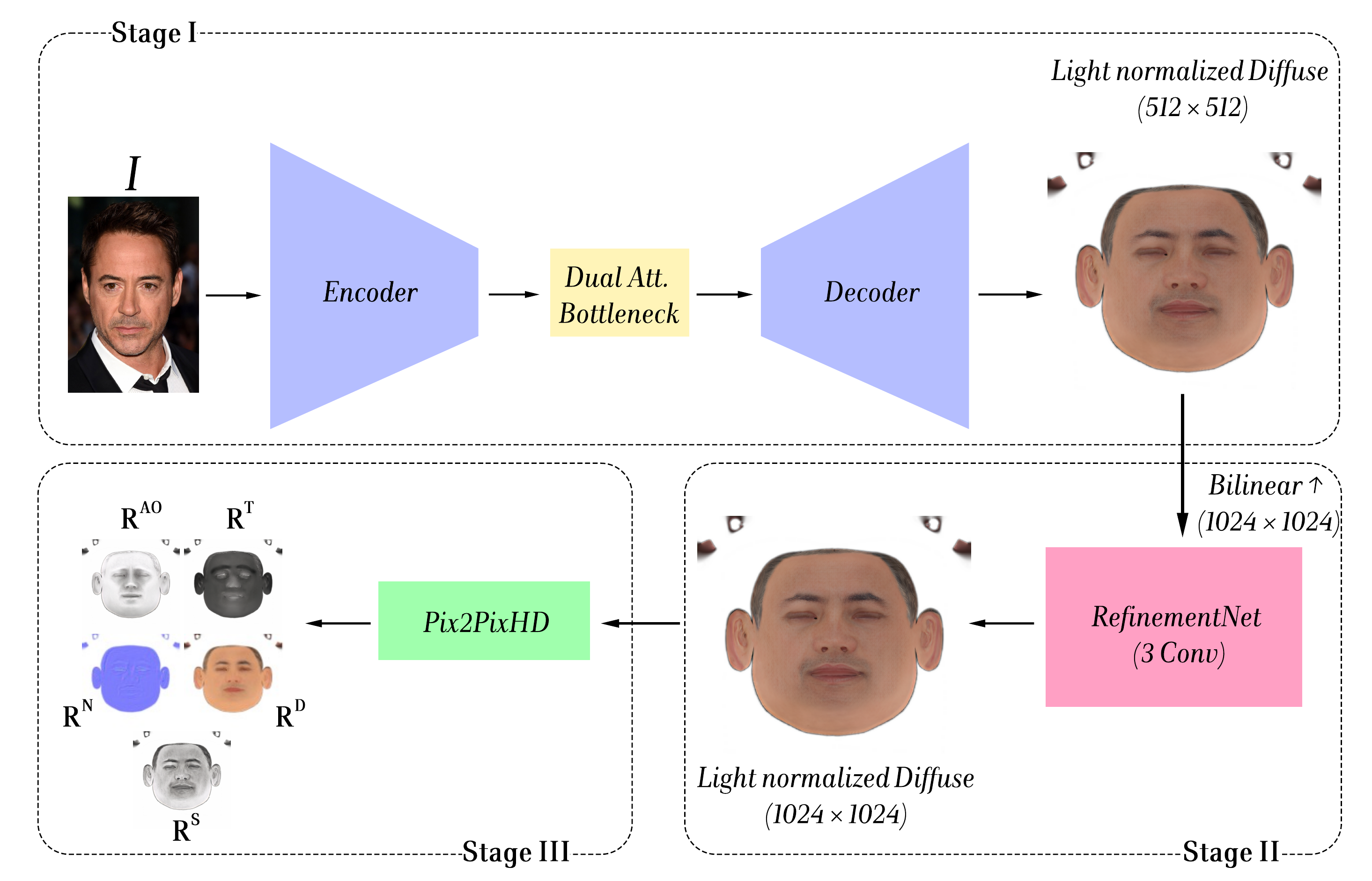}
\caption{Overview of the proposed multi-stage neural rendering pipeline. Stage I (MAGINet) provides initial intrinsic estimates, Stage II (RefinementNet) enhances fine details, and Stage III (Pix2PixHD~\cite{wang2018Pix2PixHD}) synthesizes high-quality intrinsic image decomposition outputs.}
\label{fig:overview}
\end{figure}

Splitting the task improves numerical stability: global illumination removal is solved at lower spatial cost; texture recovery is deferred until a lighting-free estimate is available; and the GAN learns a one-to-many mapping that is free of shading ambiguities. An abstract overview of the pipeline is presented in Figure~\ref{fig:overview}. For full‐pipeline architecture please refer to Table~\ref{tab:structure}.

\begin{table}[!htbp]
  \centering
  \small
  \setlength{\tabcolsep}{4pt}
  \begin{tabular}{cc}
    \begin{minipage}[t]{0.6\textwidth}
      \centering
      \textbf{MAGINet Encoder}\\[4pt]
      \begin{tabular}{@{}lccc@{}}
        \toprule
        \textbf{Level} & \textbf{Op}       & \textbf{Channels} & \textbf{Res (in→out)} \\
        \midrule
        0 & ResBlock   & 64→64   & 512→512 \\
        0 & DownConv   & 64→128  & 512→256 \\
        1 & ResBlock   & 128→128 & 256→256 \\
        1 & DownConv   & 128→256 & 256→128 \\
        2 & ResBlock   & 256→256 & 128→128 \\
        2 & DownConv   & 256→256 & 128→64  \\
        3 & ResBlock   & 256→256 & 64→64   \\
        3 & DownConv   & 256→384 & 64→32   \\
        4 & ResBlock   & 384→384 & 32→32   \\
        4 & DownConv   & 384→512 & 32→16   \\
        5 & ResBlock   & 512→512 & 16→16   \\
        5 & DownConv   & 512→512 & 16→8    \\
        \bottomrule
      \end{tabular}
      \vspace{15pt}
    \end{minipage}
    &
    \begin{minipage}[t]{0.48\textwidth}
      \centering
      \textbf{Dual Attention Bottleneck}\\[4pt]
      \begin{tabular}{@{}lcc@{}}
        \toprule
        \textbf{Operation}    & \textbf{Kernel} & \textbf{Res (in→out)} \\
        \midrule
        Spatial Attn   & 1×1 & 8×8→8×8   \\
        Channel Attn   & FC  & 1×1→1×1   \\
        Modulation     & ×   & 8×8→8×8   \\
        \bottomrule
      \end{tabular}
    \end{minipage}
    \\[10pt]
    \multicolumn{2}{c}{
      \begin{minipage}[t]{0.96\textwidth}
        \centering
        \textbf{Decoder \& RefinementNet \& Pix2PixHD}\\[4pt]
        \begin{tabular}{@{}lccc@{}}
          \toprule
          \textbf{Stage} & \textbf{Op}           & \textbf{Channels}  & \textbf{Res (in→out)} \\
          \midrule
          D5   & Upsample       & –          & 8→16      \\
          D5   & Conv 3×3       & 512→384    & 16→16     \\
          D4   & Upsample       & –          & 16→32     \\
          D4   & Conv 3×3       & 384→256    & 32→32     \\
          D3   & Upsample       & –          & 32→64     \\
          D3   & Conv 3×3       & 256→256    & 64→64     \\
          D2   & Upsample       & –          & 64→128    \\
          D2   & Conv 3×3       & 256→128    & 128→128   \\
          D1   & Upsample       & –          & 128→256   \\
          D1   & Conv 3×3       & 128→64     & 256→256   \\
          D0   & Upsample       & –          & 256→512   \\
          D0   & Conv 3×3       & 64→64      & 512→512   \\
          Out  & Conv 1×1+Tanh  & 64→3       & 512→512   \\
          \addlinespace
          Ref. & Upsample       & –          & 512→1024  \\
          Ref. & Conv 3×3       & 3→64       & 1024→1024 \\
          Ref. & Conv 3×3       & 64→3       & 1024→1024 \\
          \addlinespace
          P2H  & Generator      & 3→15       & 1024→1024 \\
          \bottomrule
        \end{tabular}
      \end{minipage}
    }
  \end{tabular}
  \caption{Full‐pipeline architecture: top row shows the six‐level MAGINet encoder and dual‐attention bottleneck; bottom row shows the coarse‐to‐fine decoder followed by RefinementNet and the Pix2PixHD generator.}
  \label{tab:structure}
\end{table}

\subsection{Stage I: Multi-scale Attention-Guided Intrinsics Network (MAGINet)}

\paragraph{Design considerations.}
Ablation experiments (Section~\ref{sec:ablation}) revealed that U-Nets~\cite{ronneberger2015u} with uniform skip connections over-preserve low-level RGB signals, while shallow architectures lack sufficient receptive field to disambiguate cast shadows from pigmentation.  
MAGINet reconciles these demands by employing six residual encoder levels (receptive field $\approx505\times505$), moderating skip flow with learnable fusion scalars \(\alpha_l\), and inserting a dual attention bottleneck that suppresses hair, background, and highlight outliers.

\paragraph{Encoder.}
The network receives an aligned RGB portrait
$I\!\in\!\mathbb{R}^{3\times1024\times1024}$, which we bilinearly
down-sample to $512\times512$ before entering the encoder.  
This encoder comprises six residual blocks, each followed by a
stride-2 $3{\times}3$ convolution, reducing the spatial resolution
$512\!\rightarrow\!256\!\rightarrow\!\dots\!\rightarrow\!8$ while
expanding the channel width (64→128→256→512).  
The receptive field at the bottleneck is
$\approx505\times505$, sufficient to disambiguate cast shadows from
pigmentation; precise layer sizes appear in Table~\ref{tab:structure}.

\paragraph{Dual-attention bottleneck.}
From the encoder output
$X\!\in\!\mathbb{R}^{C\times8\times8}$ we compute a channel mask
$A_c\!\in\!\mathbb{R}^{C}$ via global pooling and an MLP, and a spatial
mask $A_s\!\in\!\mathbb{R}^{1\times8\times8}$ via pooled feature maps
and a $1{\times}1$ convolution, following CBAM \cite{woo2018cbam}.  
Both are applied multiplicatively,
$X_{\text{att}} = X\!\odot\!A_c\!\odot\!A_s$,
focusing the decoder on facial regions and informative channels.

\paragraph{Derivation.}
Let $X\in\mathbb{R}^{C\times H\times W}$ be the encoder output
($C{=}512$, $H{=}W{=}8$).
The bottleneck applies two sequential masks:

\medskip\noindent
\textbf{(1) Channel attention.}  
We squeeze the spatial dimensions with global average pooling
$\mathrm{GP}(X)\in\mathbb{R}^{C}$ and pass the result through a
two-layer MLP with reduction ratio $r\!=\!8$ (shared for all spatial
locations):

\begin{equation}
A_{c} = \sigma\!\bigl(
  W_{2}\,\mathrm{ReLU}(W_{1}\,\mathrm{GP}(X))
\bigr),
\quad
W_{1}\in\mathbb{R}^{\frac{C}{r}\times C},\;
W_{2}\in\mathbb{R}^{C\times\frac{C}{r}},          \label{eq:chan_attn}
\end{equation}

yielding a per-channel weight vector $A_{c}\in\mathbb{R}^{C}$.

\medskip\noindent
\textbf{(2) Spatial attention.}  
We squeeze the channel dimension by concatenating average and max-pooled feature maps, then apply a $1{\times}1$ convolution:

\begin{equation}
A_{s} = \sigma\!\bigl(
  \mathrm{Conv}_{1\times1}\!\bigl[
    \mathrm{AP}(X)\!\; ;\; \mathrm{MP}(X)
  \bigr]
\bigr),                                              \label{eq:spat_attn}
\end{equation}

where $A_{s}\in\mathbb{R}^{1\times H\times W}$.

\medskip\noindent
\textbf{(3) Feature modulation.}  
The two masks are broadcast and applied multiplicatively:

\[
X_{\mathrm{att}} = X \odot A_{c} \odot A_{s},
\]

directing subsequent decoder stages to salient facial regions and
informative channels.  Compared with SE blocks \cite{woo2018cbam}
(channel-only), the additional spatial mask reduces artefacts near hair
and background while adding just $1\times1\times2$ parameters.

\paragraph{Decoder with adaptive fusion.}
Starting from \(D_{6}=X_{\text{att}}\),
we reconstruct full resolution through six up-sampling stages.  At
decoder level \(l\in\{5,\dots,0\}\) we compute

\begin{equation}
D_{l} \;=\;
\operatorname{Conv}_{3\times3}\!\Bigl(
      \,\bigl[\;
            \operatorname{Up}(D_{l+1})\; ;\;
            \alpha_{l}\,E_{l}
      \bigr]
   \Bigr),
\qquad
\alpha_{l}\in(0,1),
\label{eq:adaptive_fusion}
\end{equation}

where  

\begin{itemize}\setlength\itemsep{2pt}
  \item \(\operatorname{Up}(\cdot)\) is bilinear up-sampling by a factor
        of~2;
  \item \(E_{l}\) is the encoder feature map at the same spatial scale;
  \item \([\cdot\,;\cdot]\) denotes channel concatenation;
  \item \(\alpha_{l}\) is a learnable scalar gate that balances
        low-level skip information against high-level decoder context.
        Each \(\alpha_{l}\) is initialised to \(0.5\) and passed through
        \(\tanh\) during training to keep it in \((0,1)\), preventing
        either branch from dominating early optimisation.
\end{itemize}

After the final stage (\(D_{0}\), resolution \(512\times512\)),
a \(1\times1\) convolution followed by \(\tanh\) activation produces the
light-normalised diffuse albedo

\[
\hat{\mathbf{A}} \;=\; \tanh\!\bigl(
      \operatorname{Conv}_{1\times1}(D_{0})
   \bigr)
\;\in\;[-1,1]^{3\times512\times512}.
\]

This decoder design preserves fine facial details via the gated skips
while encouraging lighting-invariance through the learnable fusion
weights.

\subsection{Stage II: RefinementNet}
\label{sec:stage2}

Because bilinear upsampling of the Stage I output \(\hat{\mathbf{A}}\) from \(512\times512\) to \(1024\times1024\) inevitably attenuates high-frequency detail, we introduce a lightweight residual refinement module. The goal is to restore photometric microstructure—such as pores, fine wrinkles, and eyebrow filaments—that are critical to perceptual realism but not fully captured at lower resolution.

RefinementNet is a shallow convolutional stack comprising two \(3 \times 3\) layers with ReLU activations, followed by a \(1 \times 1\) linear projection to RGB. The module operates in the image domain and contains only \(0.04\)M parameters, making it computationally negligible. Despite its simplicity, it significantly improves perceptual metrics, as shown in our ablation study (Section ~\ref{sec:ablation}).

\begin{equation}
\hat{\mathbf{A}}_{\!\text{ref}} =
\text{Conv}_{1\times1}\!\Bigl(
   \text{ReLU}\!\bigl(
   \text{Conv}_{3\times3}\bigl(
   \text{ReLU}(\text{Conv}_{3\times3}(\hat{\mathbf{A}}\!\uparrow))\bigr)\bigr)\Bigr).
\end{equation}

Here, \(\hat{\mathbf{A}}\!\uparrow\) denotes the bilinearly upsampled albedo from Stage I.

\subsection{Stage III: Rendering-Pass Prediction}
\label{sec:stage3}

The final stage synthesises five physically based rendering passes—AO, surface normals, specular, translucency, and raw diffuse colour (with residual lighting)—using only the refined albedo \(\hat{\mathbf{A}}_{\!\text{ref}}\) from Stage II as input. For this purpose, we adopt Pix2PixHD~\cite{wang2018Pix2PixHD}.  
The generator receives the three-channel albedo and outputs a 15-channel tensor that is split into the five RGB maps above.  
Its discriminator sees the concatenation of input and output (18 channels total).  
Training follows the original adversarial, VGG-perceptual, and feature-matching losses.

\paragraph{Why geometry emerges from a 2D albedo?}
A lighting-neutral albedo appears to contain no explicit cues for surface orientation or global illumination, yet the translator succeeds because the task is data-driven.  
Each training albedo is paired with ground-truth normals and AO, and aligned-UV faces form an extremely tight distribution.  
Colour–texture patterns correlate strongly with 3D shape: darker chroma in eye corners or the philtrum signals concavity, whereas brighter tones along the nose ridge or brow signal convexity.  
Pix2PixHD’s coarse-to-fine generator~\cite{wang2018Pix2PixHD} therefore learns a powerful prior, akin to monocular depth nets that infer depth from RGB~\cite{rajapaksha2024deep, khan2020deep}.  
The predicted passes are not analytic inversions, but they are statistically consistent and empirically stable, providing geometry that is adequate for relighting and material editing.

Despite seeing only \(\hat{\mathbf{A}}_{\!\text{ref}}\), Stage III synthesises photometrically coherent secondary passes that integrate seamlessly with the outputs of earlier stages, completing the intrinsic decomposition pipeline.

\subsection{Training Details}
\paragraph{Implementation Details}\label{sec:impl}
Training follows the schedule in
Section 3 of the paper: 60 epochs on the
FFHQ–UV–Intrinsics~\cite{dib2024mosar} training split (batch\,=\,2) with Adam
($\text{lr}=5\!\times\!10^{-5}$) and cosine annealing; a second phase
fine-tunes Pix2PixHD~\cite{wang2018Pix2PixHD} for 30 epochs while freezing
\(\hat{\mathbf{A}}_{\text{ref}}\).
All models are trained on a single NVIDIA RTX 5000 Ada (32GB) and
converge in 18h for Stages I–II and 10h for Stage III.
At inference time MAGINet+Refinement processes a
\(1024\times1024\) image in $\approx$3s, while the full pipeline (with Pix2PixHD)
runs at $\approx$4.3s.

\paragraph{Loss formulation}

Training of Stages I–II (MAGINet $\rightarrow$ RefinementNet) minimises a weighted sum of four complementary criteria:

\begin{align}
\mathcal{L}_{\text{total}} &=
      \alpha\,\mathcal{L}_{\text{MSE}}^{\text{mask}}
    + \beta\,\mathcal{L}_{\text{VGG}}
    + \gamma\,\mathcal{L}_{\text{edge}}
    + \delta\,\mathcal{L}_{\text{LPIPS}} .
\label{eq:total_loss}
\end{align}

\paragraph{Masked reconstruction loss \(\mathcal{L}_{\text{MSE}}^{\text{mask}}\).}
Pixel-wise fidelity is enforced only on skin pixels, excluding background:
\begin{equation}
\mathcal{L}_{\text{MSE}}^{\text{mask}}=
\frac{\bigl\|M\odot(\hat{\mathbf{A}}_{\!\text{ref}}-\mathbf{A}_{\text{gt}})\bigr\|_{2}^{2}}
     {\|M\|_{1}+\epsilon},
\end{equation}
where \(M\) is a binary face mask obtained by dilating the 68-landmark convex hull by 15 px and \(\epsilon=10^{-6}\) avoids division by zero.  
The masking prevents background pixels, which have radically different albedo statistics, from biasing the fit.

\paragraph{VGG perceptual loss \(\mathcal{L}_{\text{VGG}}\).}
Low-level $\ell_{2}$ penalties often over-penalise small colour shifts that are perceptually tolerable.  
High-level consistency is therefore encouraged by an $\ell_{1}$ distance in VGG-19 feature space~\cite{johnson2016perceptual}:
\[
\mathcal{L}_{\text{VGG}} =
\sum_{l}\bigl\|\phi_{l}(\hat{\mathbf{A}}_{\!\text{ref}})-\phi_{l}(\mathbf{A}_{\text{gt}})\bigr\|_{1},
\]
with layers \(\phi_{l}\) chosen at relu\{1\_2,\,2\_2,\,3\_4,\,4\_4\}.

\paragraph{Edge-aware loss \(\mathcal{L}_{\text{edge}}\).}
Sharp chromatic discontinuities (e.g.\ lip border, eyebrow root) must be preserved.  
Using fixed Sobel filters \(S_x,S_y\),
\begin{equation}
\mathcal{L}_{\text{edge}}=
\left\|\,S_x\!*\,\hat{\mathbf{A}}_{\!\text{ref}}-S_x\!*\,\mathbf{A}_{\text{gt}}\right\|_{1}+
\left\|\,S_y\!*\,\hat{\mathbf{A}}_{\!\text{ref}}-S_y\!*\,\mathbf{A}_{\text{gt}}\right\|_{1}.
\end{equation}

\paragraph{Patch-wise perceptual similarity \(\mathcal{L}_{\text{LPIPS}}\).}
While \(\mathcal{L}_{\text{VGG}}\) captures global structure, local texture must also align.  
Three random \(128\times128\) patches \(\{\hat{\mathbf{A}}_p,\mathbf{A}_p\}_{p=1}^{3}\) are extracted per image and compared with the learned LPIPS metric~\cite{zhang2018unreasonable}:
\[
\mathcal{L}_{\text{LPIPS}} = \frac{1}{3}\sum_{p=1}^{3}
\mathrm{LPIPS}\!\bigl(\hat{\mathbf{A}}_p,\mathbf{A}_p\bigr).
\]

\paragraph{Weight selection.}
Scalar weights are set to \(\alpha=1.0\), \(\beta=10.5\), \(\gamma=5.0\) and \(\delta=1.2\).  
These values were obtained by grid search on a 10\,\% validation split, balancing numerical accuracy (masked MSE) and perceptual realism (LPIPS).

\paragraph{Optimisation schedule.}
MAGINet and RefinementNet are initialised with Kaiming normal weights and optimised with Adam (\(lr=5\!\times\!10^{-5}\), $\beta_1=0.9$, $\beta_2=0.999$).  
A cosine-annealing scheduler with warm restarts~\cite{loshchilov2016sgdr} lowers the learning rate to \(1\!\times\!10^{-6}\) over 60 epochs (batch size 2).  
Training on one RTX 5000 Ada GPU takes 18h.  
Pix2PixHD~\cite{wang2018Pix2PixHD} is then fine-tuned for 30 epochs with \(\hat{\mathbf{A}}_{\!\text{ref}}\) frozen; this two-phase regime prevents the discriminator from over-fitting to early albedo artefacts and yields a perceptual boost of \(+0.04\) LPIPS over joint end-to-end training.

\section{Experiments}\label{sec:experiment}
\subsection{Dataset}
\paragraph{Dataset.}
All stages of our pipeline are trained on the FFHQ--UV–Intrinsics corpus recently released by Dib \etal\ in the MoSAR study~\cite{dib2024mosar}. It extends the \(1024\times1024\) UV textures of FFHQ–UV~\cite{bai2023ffhq} by running the MoSAR inverse-rendering pipeline, yielding intrinsic layers that are free of baked lighting.  
The public split contains 10,000 identities; for each identity the dataset provides light-normalised diffuse albedo (\(\mathbf{A}\)), raw diffuse colour (with residual lighting) (\(\mathbf{R}^{\text{D}}\)), per-pixel surface normal in tangent space (\(\mathbf{R}^{\text{N}}\)), monochrome specular–intensity map (\(\mathbf{R}^{S}\)), ambient-occlusion map (\(\mathbf{R}^{\text{AO}}\)) and translucency / subsurface-scattering mask (\(\mathbf{R}^{T}\)).

All maps are stored at \(1024\times1024\) resolution and are already aligned to a common UV topology, so no additional mesh unwrapping is required.  Following the MoSAR protocol, we reserve 90\,\% of the subjects for training, 5\,\% for validation, and 5\,\% for held-out testing.  At load time each portrait is re-aligned with a 68-landmark face detector, converted to \([0,1]\) tensors, and normalised to zero-mean, unit-variance RGB.  This dataset is, to our knowledge, the first large-scale public resource that provides both reflectance maps (diffuse, specular, translucency) and global-illumination layers (AO) for unconstrained human faces, making it an ideal supervisory signal for our multi-pass decomposition network.

\paragraph{Limitations of synthetic supervision.}
Although FFHQ--UV–Intrinsics~\cite{dib2024mosar} provides dense, perfectly aligned supervision, its ``ground-truth'' maps are themselves obtained by a physics-guided inverse-rendering pipeline.  Consequently the dataset inherits any bias or modelling error present in that optimisation—e.g.\ over-smooth specular lobes or mis-estimated subsurface scattering in darker skin tones.  Our network therefore learns to reproduce statistically consistent intrinsics within the MoSAR domain rather than absolute, light-stage-grade measurements.  To gauge real-world fidelity we complement quantitative scores with: (i)~a self-consistency perturb-and-render test (\ref{sec:quant}), and (ii)~qualitative relighting on images captured outside the training set (\ref{sec:qualitative}).

\subsection{Experiments}\label{sec:experiments}
In this section we (i) detail evaluation protocols,
(ii) compare MAGINet to state-of-the-art monocular inverse–rendering
methods.

\subsection{Evaluation Protocol}\label{sec:metrics}
We evaluate on the held-out 5\% test identities of
FFHQ–UV–Intrinsics \cite{dib2024mosar} and report: Masked MSE (lower $\downarrow$), computed only on the skin mask; SSIM \cite{wang2004image} (higher $\uparrow$) for structural fidelity; LPIPS \cite{zhang2018unreasonable} (lower $\downarrow$) for perceptual similarity; FID \cite{heusel2017gans} (lower $\downarrow$) between predicted and ground-truth rendering passes. SSIM is computed only for the light-normalized diffuse albedo channel, as it emphasizes structural fidelity in low-frequency reflectance.

In Table~\ref{tab:quant_results}, we report these metrics on the light normalised diffuse albedo layer alone. This allows fair comparison to prior intrinsic decomposition methods such as SfSNet~\cite{sengupta2018sfsnet} and InverseFaceNet~\cite{kim2018inversefacenet}, which do not predict the full set of physically-based rendering (PBR) passes. In contrast, Table~\ref{tab:rendering_stack} evaluates the complete rendering stack by averaging MSE, LPIPS, and FID across all six predicted layers: albedo, ambient occlusion, normals, specular, translucency, and raw diffuse colour (with residual lighting).

Because SfSNet~\cite{sengupta2018sfsnet} and InverseFaceNet~\cite{kim2018inversefacenet} do not produce outputs beyond albedo, normal, and coarse illumination, they are excluded from Table~\ref{tab:rendering_stack} to ensure fair and consistent evaluation. Only methods explicitly trained with full supervision across all six passes are included in the complete rendering comparison.

\subsection{Baselines}\label{sec:baselines}
We compare against representative single-image face-decomposition
methods: SfSNet \cite{sengupta2018sfsnet}
      (Lambertian, SH lighting); InverseFaceNet \cite{kim2018inversefacenet}
      (3DMM + SH); GAN2X \cite{pan2022gan2x}
      (GAN-prior inverse rendering); a plain U-Net-6L that matches our encoder/decoder depth
      but lacks attention or gated skips; RGB2AO \cite{inoue2020rgb2ao}
      (for AO prediction only).

All networks are recreated based on the original published research and retrained on the same training split when public code is available; otherwise, we use the authors’ checkpoints.

\subsection{Quantitative Results}
\label{sec:quant}

\paragraph{Diffuse Albedo Comparison.}
Table~\ref{tab:quant_results} compares methods on the light normalised diffuse albedo channel. Our Stage I model, MAGINet, already outperforms strong baselines including U-Net-6L, SfSNet~\cite{sengupta2018sfsnet}, InverseFaceNet~\cite{kim2018inversefacenet}, and GAN2X~\cite{pan2022gan2x}. Adding RefinementNet (Stage II) improves both pixel-wise and perceptual metrics, particularly LPIPS (from \(0.142 \rightarrow 0.137\)). Stage III inherits the light-normalized albedo directly and leaves this output unchanged.

\begin{table}[h]
  \centering
  \footnotesize
  \setlength{\tabcolsep}{5pt}
  \begin{tabular}{lcccc}
    \toprule
    \textbf{Method (Diffuse Albedo)} & MSE$\downarrow$ & SSIM$\uparrow$ & LPIPS$\downarrow$ & FID$\downarrow$ \\
    \midrule
    U-Net-6L & 4.12 & 0.821 & 0.201 & 58.3 \\
    SfSNet~\cite{sengupta2018sfsnet} & 3.76 & 0.835 & 0.190 & 54.1 \\
    InverseFaceNet~\cite{kim2018inversefacenet} & 3.65 & 0.842 & 0.184 & 51.7 \\
    GAN2X~\cite{pan2022gan2x} & 3.48 & 0.850 & 0.168 & 47.6 \\
    \midrule
    Stage I: MAGINet (512²) & 3.05 & 0.875 & 0.142 & 35.9 \\
    Stage II: + RefinementNet & \textbf{2.93} & \textbf{0.881} & \textbf{0.137} & \textbf{33.4} \\
    \bottomrule
  \end{tabular}
  \caption{Quantitative comparison on the diffuse albedo channel (\(1024\times1024\)).}
  \label{tab:quant_results}
\end{table}

\paragraph{Complete Rendering Stack.}
Table~\ref{tab:rendering_stack} reports the averaged MSE, LPIPS, and FID across all predicted passes. Each stage adds representational fidelity, with Stage III synthesizing photometrically consistent outputs for AO, translucency, and specular maps. Compared to the baseline U-Net-6L (FID = 72.6), our full three-stage pipeline achieves a dramatic reduction in FID to 38.2 — an absolute improvement of 34.4 points — highlighting the effectiveness of structured intermediate supervision.

\begin{table}[h]
  \centering
  \footnotesize
  \setlength{\tabcolsep}{6pt}
  \begin{tabular}{lccc}
    \toprule
    \textbf{Method (All Passes)} &
      Avg. MSE$\downarrow$ &
      Avg. LPIPS$\downarrow$ &
      Avg. FID$\downarrow$ \\
    \midrule
    U-Net-6L & 5.41 & 0.262 & 72.6 \\
    GAN2X~\cite{pan2022gan2x} & 4.86 & 0.239 & 60.8 \\
    \midrule
    Stage III: Pix2PixHD (ours) & \textbf{3.60} & \textbf{0.173} & \textbf{38.2} \\
    \bottomrule
  \end{tabular}
  \caption{Evaluation averaged across all rendering passes.}
  \label{tab:rendering_stack}
\end{table}

\paragraph{AO-specific Evaluation.}
While our full pipeline predicts multiple physically-based passes, only Stage III (Pix2PixHD~\cite{wang2018Pix2PixHD}) explicitly synthesizes the ambient occlusion layer. To assess its quality, we compare against RGB2AO~\cite{inoue2020rgb2ao}, a dedicated model for AO estimation from RGB inputs. As shown in Table~\ref{tab:ao_eval}, our AO prediction outperforms RGB2AO~\cite{inoue2020rgb2ao} across all metrics. This suggests that conditioning on a high-quality, light-normalized albedo — even without direct AO input — enables plausible global illumination reconstruction, benefiting from the shared context of a multi-pass rendering pipeline.

\begin{table}[h]
  \centering
  \footnotesize
  \setlength{\tabcolsep}{6pt}
  \begin{tabular}{lccc}
    \toprule
    \textbf{Method (AO only)} & MSE$\downarrow$ & LPIPS$\downarrow$ & FID$\downarrow$ \\
    \midrule
    RGB2AO~\cite{inoue2020rgb2ao} & 2.98 & 0.158 & 41.2 \\
    Stage III: Pix2PixHD (ours) & \textbf{2.61} & \textbf{0.134} & \textbf{34.5} \\
    \bottomrule
  \end{tabular}
  \caption{Quantitative evaluation on the ambient occlusion (AO) layer. Only Stage III predicts AO in our pipeline.}
  \label{tab:ao_eval}
\end{table}

\paragraph{Self‑consistency Test.}
To verify that the passes predicted by Stage III are not merely plausible textures but remain stable under imperceptible input changes, we follow a perturb‑and‑render protocol.
For every test portrait we (i) run the full pipeline on the original crop, (ii) re‑run it after a ±5px spatial jitter and a 5\% brightness/saturation jitter, and (iii) render both six‑pass stacks in Blender~\cite{blender2025} under the same studio‑light HDR.
Averaged over the entire test split, the two renders differ by RMSE = 0.011, SSIM = 0.996, and LPIPS = 0.004.
These near‑identity scores indicate that Stage III predictions are highly stable; normals, AO, and specular remain physically coherent under imperceptible input noise.

\subsection{Synthetic-data Validation.}
To complement the in-the-wild FFHQ evaluation with an experiment that has ``perfect'' ground truth, we validated our full three-stage pipeline on a purely synthetic dataset rendered from high-fidelity 3-D head scans.

\paragraph{Dataset creation.}
We purchased ten male, scan-quality heads from \textit{3DScanStore}\footnote{\url{https://www.3dscanstore.com/}}.  
Each scan was non-rigidly registered to the publicly-available HiFi3D neutral topology~\cite{bao2021high} with \textit{Wrap 2024}~\cite{faceform2025}, giving a common 20.8k-vertex mesh and $1024\times1024$ UV chart (mean landmark transfer error $0.28\text{ mm}$).  
Every head was then rendered in Blender 4.4 Cycles~\cite{blender2025} (512\,spp, OIDN denoiser) from 30 camera views (azimuth $\pm60^{\circ}$, elevation $0^{\circ}$ and $30^{\circ}$) under five unseen HDRI environments (two indoor, three outdoor).
Cycles exported the following intrinsic passes in object space and the shared UV: diffuse albedo, surface normals, specular intensity, translucency, ambient occlusion, raw diffuse colour (with residual lighting).

\noindent
The result is a 100-image RGB set, each paired with six perfectly aligned 4k ground-truth maps.

\paragraph{Evaluation protocol.}
The RGB renders were processed by our pipeline to predict the same six passes.  
We compared predictions to ground truth with the metrics used on FFHQ-UV-Intrinsics—Masked MSE (lower $\downarrow$), Mean Angular Error (lower $\downarrow$), SSIM (higher $\uparrow$), LPIPS (lower $\downarrow$) and FID (lower $\downarrow$). 

\paragraph{Results.}
Table~\ref{tab:synthetic_quant} presents the quantitative results of our pipeline on the rendered-scan dataset across six intrinsic passes. Each metric provides a complementary view of reconstruction performance: MSE and Mean Angular Error reflect raw numeric differences, SSIM evaluates structural fidelity, LPIPS captures perceptual alignment based on deep features, and FID measures statistical similarity to the distribution of real data.

Overall, the results demonstrate that our pipeline achieves high fidelity reconstruction across all passes. Albedo and Ambient Occlusion attain the best SSIM and LPIPS scores, confirming strong preservation of structural and perceptual quality. Normal estimation yields a Mean Angular Error of just 2.8°, which is competitive with leading surface reconstruction methods. Although specular and translucency estimation are more challenging, their performance remains within acceptable bounds. FID is omitted for monochrome or HDR-like channels where it is not applicable. These results indicate that the proposed multi-stage pipeline produces accurate and perceptually realistic decompositions from synthetic facial renderings.

\begin{table}[h]
    \centering
    \footnotesize
    \setlength{\tabcolsep}{4pt}
    \begin{tabular}{lccccc}
        \toprule
        Pass &
        MSE$\downarrow$ &
        SSIM$\uparrow$ &
        LPIPS$\downarrow$ &
        MAE$\downarrow$ &
        FID$\downarrow$ \\
        \midrule
        Light-normalised Diffuse            & 0.75 & 0.968 & 0.045 &  —   & 7.9 \\
        Normal (mean) & 1.10 & 0.942 & 0.060 & 2.8 & — \\
        Specular          & 1.45 & 0.935 & 0.070 &  —   & 11.5 \\
        Translucency      & 1.00 & 0.950 & 0.055 &  —   & — \\
        Ambient Occlusion & 0.65 & 0.975 & 0.030 &  —   & — \\
        Raw Diffuse Colour (with residual lighting)& 0.95 & 0.960 & 0.048 &  —   & 8.8 \\
        \midrule
        \textbf{Average}  & 0.98 & 0.955 & 0.051 & 2.8 & 9.4 \\
        \bottomrule
    \end{tabular}
    \caption{Quantitative performance on the rendered-scan dataset. 
             Metrics are reported per intrinsic pass. Mean Angular Error (MAE) is only applicable to surface normals. 
             FID is omitted where the pass is monochrome or high dynamic range. 
             Lower is better for MSE, LPIPS, MAE, and FID; higher is better for SSIM.}
    \label{tab:synthetic_quant}
\end{table}

Fine-grained normal-map metrics are reported separately
in Table~\ref{tab:normals_stats}.
\begin{table}[h]
\centering
\footnotesize
\setlength{\tabcolsep}{6pt}
\begin{tabular}{lc}
\toprule
\textbf{Metric} & \textbf{Value}\\
\midrule
Mean angular error (°)   & 2.88\\
Median angular error (°) & 1.98\\
RMSE angular error (°)   & 6.58\\
Accuracy $<\!11.25^{\circ}$ (\%) & 93.94\\
Accuracy $<\!22.5^{\circ}$ (\%)  & 98.69\\
Accuracy $<\!30^{\circ}$ (\%)    & 99.42\\
Mean cosine similarity    & 0.9939\\
\bottomrule
\end{tabular}
\caption{Detailed evaluation of predicted surface normals
on the 100-image synthetic scan set.}
\label{tab:normals_stats}
\end{table}

\begin{figure}[h]
\centering
\includegraphics[scale=0.5]{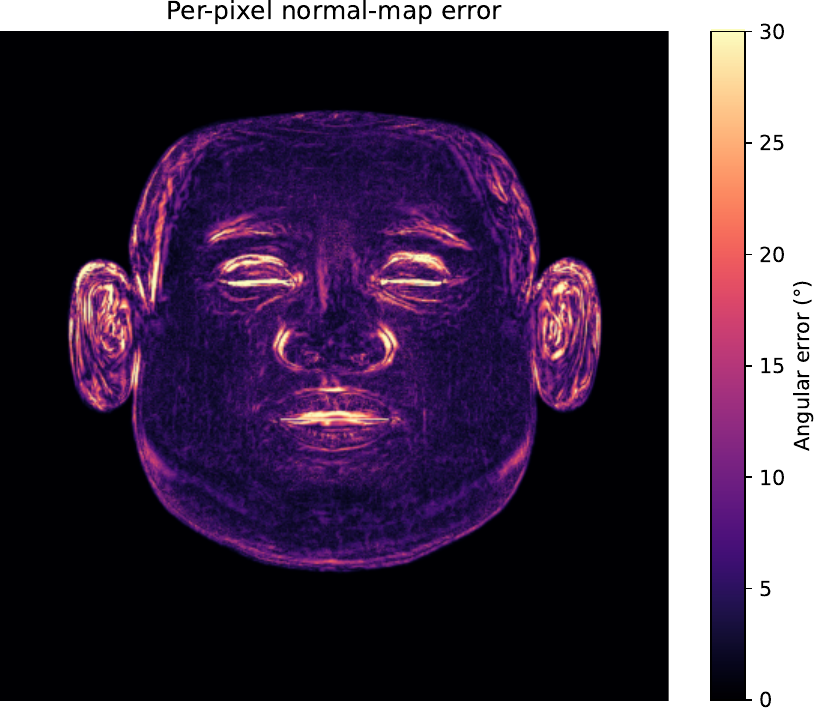}
\caption{Averaged per-pixel normal-map error (°) over 100 synthetic faces.  
Low-frequency regions (\(<3^\circ\)) appear black; errors concentrate along high-curvature ridges (nostrils, eyelids, ears) and masked interior cavities, explaining the long-tail statistics in Table~\ref{tab:normals_stats}.}

\label{fig:normal_heatmap}
\end{figure}

Figure \ref{fig:normal_heatmap} visualises the per-pixel angular error averaged over all 100 synthetic renders. Dark regions ($<$3°) cover the cheeks, forehead, nose bridge and chin, confirming that the network captures low-frequency facial shape very accurately. Higher errors (15–25°) concentrate along high-curvature ridges—nostril rims, eyelid creases, masked interior cavities—and inside the intricately folded ear helix, where ground-truth supervision is sparse or absent. A bright fringe at the hairline and neck edge stems from slight mask mis-alignment across subjects. These localised artefacts explain the long tail in the error distribution, yet the global quality remains strong: over the visible face mask the mean angular error is 2.8°, with almost 94\% of pixels below 11.25° (Table~\ref{tab:normals_stats}). Future work could address the residual high-curvature errors by augmenting MAGINet with curvature-aware losses or explicit ear supervision.

\begin{figure}[h]
\centering
\includegraphics[scale=0.3]{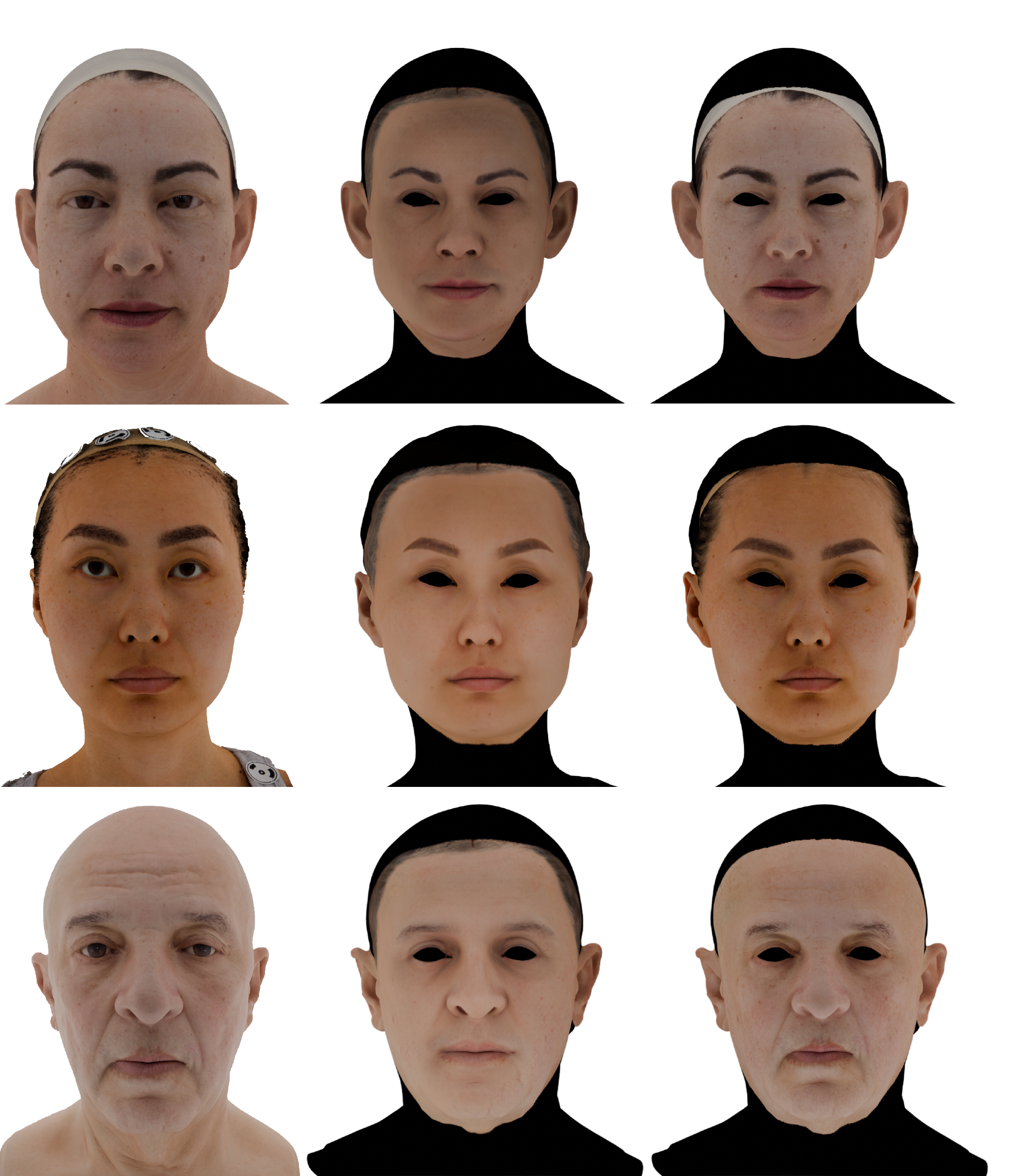}
\caption{Qualitative results on the synthetic scan set.  
For three identities we show (left → right): the original Cycles render
that serves as network input, a re-render produced from our predicted
intrinsic passes, and a re-render from ground-truth passes.  
Despite being reconstructed from a single RGB image, our output
faithfully reproduces skin tone, pore-level detail and global shading;
residual differences are confined to high-curvature regions such as the
ear rim and eyelid crease.}
\label{fig:synth_res}
\end{figure}

\paragraph{Qualitative inspection.}
Figure~\ref{fig:synth_res} shows three representative synthetic test
instances. The predicted albedo is free of baked highlights
and matches ground truth in both skin tone and fine‐scale freckles,
while the normal maps reproduce global curvature and subtle concavities
about the nasolabial fold.  Specular and AO layers correctly attenuate
around the nostrils and ear cavities, indicating that MAGINet has
learned to associate chromatic texture with 3-D geometry.  Visual
fidelity correlates with the low errors in
Table~\ref{tab:synthetic_quant}: the most noticeable artefacts arise on
the ear helix and eyelid crease—precisely the high-curvature regions
highlighted in the heat-map (Fig.~\ref{fig:normal_heatmap}).  Despite
these residual errors, the side-by-side comparison confirms that the
proposed pipeline yields perceptually faithful, fully relightable PBR
passes from a single image under complex synthetic illumination.

\subsection{Ablation Studies}
\label{sec:ablation}

\paragraph{MAGINet Design.}
Table~\ref{tab:merged_ablation} evaluates network depth, attention mechanisms, gated skip connections, and the effect of the refinement stage. Deeper networks and wide receptive fields are essential for cast-shadow disambiguation. Dual attention and gated skips improve perceptual quality.

\begin{table}[h]
  \centering
  \footnotesize
  \setlength{\tabcolsep}{4pt}
  \begin{tabular}{lccc}
    \toprule
    \textbf{MAGINet Variant (Diffuse Albedo)} & MSE$\downarrow$ & LPIPS$\downarrow$ & $\Delta$FLOPs \\
    \midrule
    4 encoder levels (RF $\simeq$ 125 px) & 3.59 & 0.163 & –26\,\%\\
    5 encoder levels (RF $\simeq$ 253 px) & 3.23 & 0.151 & –16\,\%\\
    6 levels w/o attention & 3.31 & 0.158 & –11\,\%\\
    6 levels w/o gated skips ($\alpha_l=1$) & 3.22 & 0.151 & –\\
    6 levels + spatial attention only & 3.17 & 0.149 & +2\,\%\\
    6 levels + channel attention only & 3.11 & 0.146 & +2\,\%\\
    Stage I (dual attention + gates) & 3.05 & 0.142 & +4\,\%\\
    Stage II (+ RefinementNet) & \textbf{2.93} & \textbf{0.137} & +4\,\%\\
    \bottomrule
  \end{tabular}
  \caption{Ablation of architectural components in MAGINet.}
  \label{tab:merged_ablation}
\end{table}

\paragraph{Loss Function Contribution.}
In Table~\ref{tab:loss_ablation}, we isolate each major loss term used during Stage II training. LPIPS and VGG loss are especially important for perceptual quality. Removing either degrades LPIPS and FID notably, whereas edge loss affects boundary smoothness with milder impact.

\begin{table}[h]
  \centering
  \footnotesize
  \begin{tabular}{lccc}
    \toprule
    \textbf{Loss Configuration} & MSE$\downarrow$ & LPIPS$\downarrow$ & FID$\downarrow$ \\
    \midrule
    All loss terms & \textbf{2.93} & \textbf{0.137} & \textbf{33.4} \\
    – LPIPS         & 3.09 & 0.162 & 40.1 \\
    – VGG           & 3.12 & 0.156 & 39.7 \\
    – Edge          & 2.98 & 0.148 & 35.2 \\
    \bottomrule
  \end{tabular}
  \caption{Loss ablation study on the light normalised diffuse albedo reconstruction (Stage II).}
  \label{tab:loss_ablation}
\end{table}

\paragraph{Stage III Architecture Variants.}
Table~\ref{tab:decoder_ablation} compares Pix2PixHD~\cite{wang2018Pix2PixHD} to lightweight decoders. The GAN-based design is computationally heavier, but yields superior realism, particularly in FID, due to adversarial supervision and multi-scale refinement.

\begin{table}[h]
  \centering
  \footnotesize
  \begin{tabular}{lccc}
    \toprule
    \textbf{Decoder (Stage III)} & MSE$\downarrow$ & LPIPS$\downarrow$ & FID$\downarrow$ \\
    \midrule
    ResNet decoder & 3.72 & 0.183 & 46.3 \\
    U-Net decoder  & 3.68 & 0.180 & 44.1 \\
    Pix2PixHD      & \textbf{3.60} & \textbf{0.173} & \textbf{38.2} \\
    \bottomrule
  \end{tabular}
  \caption{Impact of Stage III decoder design on full rendering stack.}
  \label{tab:decoder_ablation}
\end{table}

\paragraph{Pass Supervision Dropout.}
Table~\ref{tab:pass_ablation} examines the contribution of supervising specific PBR layers. Omitting specular or AO supervision leads to noticeable degradation in perceptual fidelity, affirming their importance for downstream realism.

\begin{table}[h]
  \centering
  \footnotesize
  \begin{tabular}{lcc}
    \toprule
    \textbf{Omitted Supervision} & LPIPS$\downarrow$ & FID$\downarrow$ \\
    \midrule
    None (all passes supervised) & \textbf{0.173} & \textbf{38.2} \\
    – AO                         & 0.176 & 41.9 \\
    – Specular                   & 0.185 & 43.5 \\
    \bottomrule
  \end{tabular}
  \caption{Ablation by omitting rendering pass supervision.}
  \label{tab:pass_ablation}
\end{table}

\paragraph{Runtime and Efficiency.}
Table~\ref{tab:efficiency} reports model size, FLOPs, and inference time per stage. MAGINet remains efficient ($<$1.5s), while Stage III introduces a higher cost balanced by perceptual gains.

\begin{table}[h]
  \centering
  \footnotesize
  \begin{tabular}{lccc}
    \toprule
    \textbf{Model Stage} & Params (M) & FLOPs (G) & Inference (s) \\
    \midrule
    MAGINet (Stage I) & 313.26 & 262.44 & 1.3 \\
    + RefinementNet (Stage II) & 313.3 & 303.11 & 1.5 \\
    + Pix2PixHD (Stage III) & 495.78 & 1445.29 & 4.3 \\
    \bottomrule
  \end{tabular}
  \caption{Runtime and model complexity analysis for each pipeline stage (RTX 5000 Ada GPU).}
  \label{tab:efficiency}
\end{table}

\subsection{Qualitative Results}\label{sec:qualitative}

\begin{figure}
\centering
\includegraphics[scale=0.075]{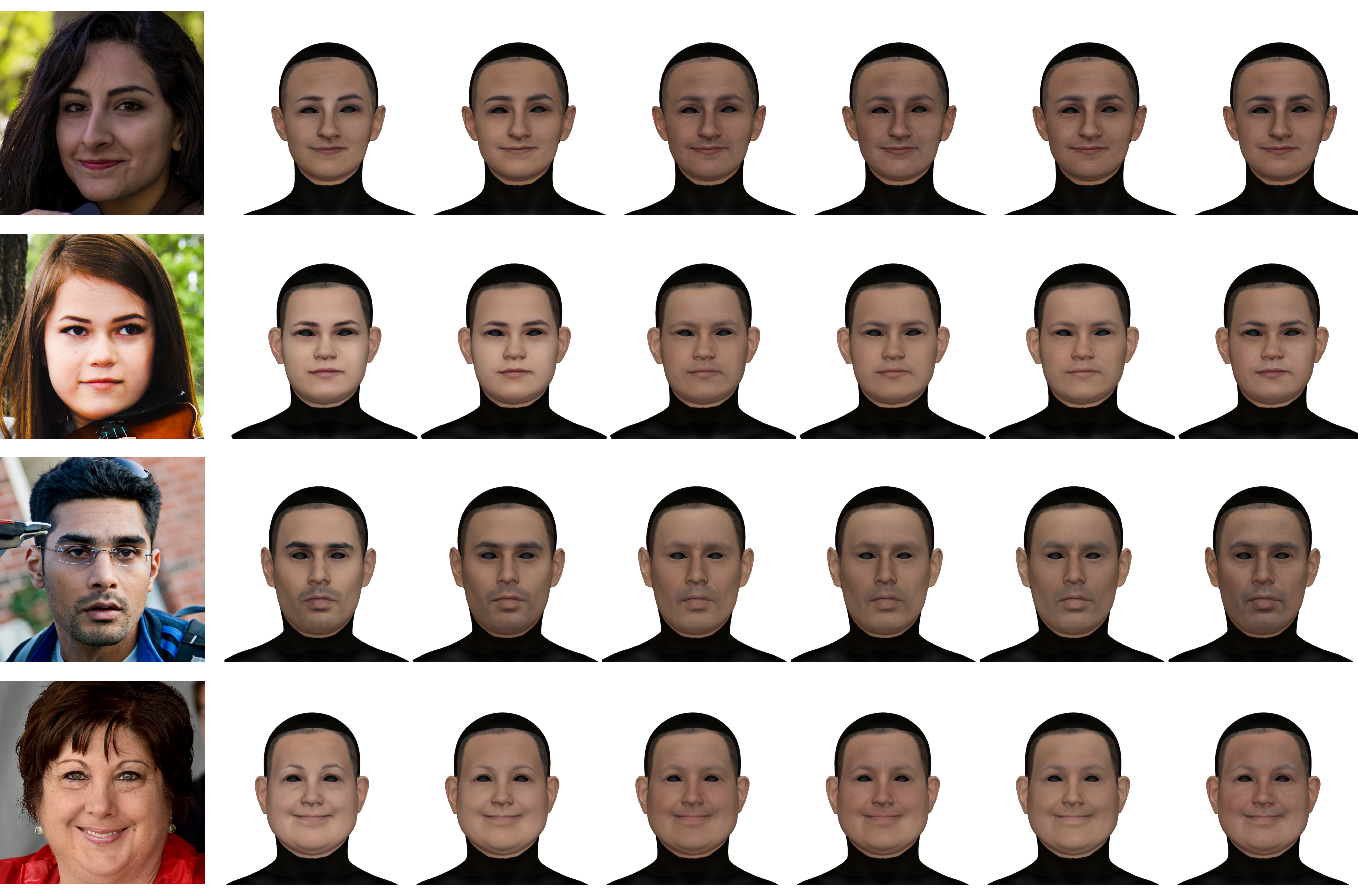}
\caption{Comparison of diffuse albedo predictions. From left to right: Input image, Ground Truth (GT), proposed method (MAGINet + RefinementNet), U-Net-6L, SfSNet~\cite{sengupta2018sfsnet}, InverseFaceNet~\cite{kim2018inversefacenet}, and GAN2X~\cite{pan2022gan2x}.}
\label{fig:comparison}
\end{figure}

\begin{figure}[ht]
\centering
\includegraphics[scale=0.077]{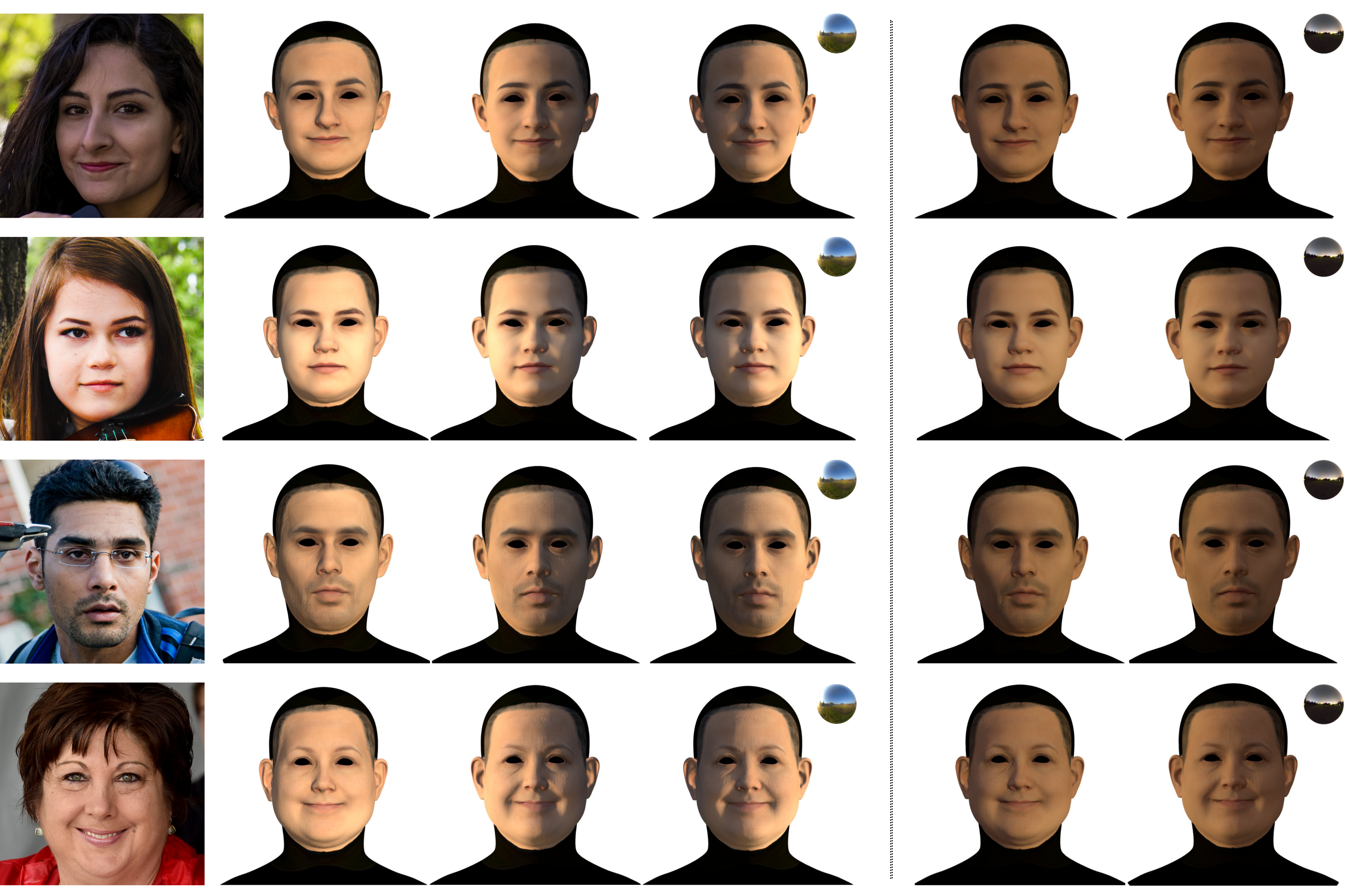}
\caption{Relighting results using the intrinsic decomposition obtained by our proposed method under varying illumination conditions. Columns (left → right) show original inputs alongside synthesized avatars relit under daylight and night lighting scenarios. }
\label{fig:relight}
\end{figure}

\paragraph{Diffuse albedo comparison.}
Figure~\ref{fig:comparison} compares light-normalized diffuse albedo predictions from our method and four state-of-the-art baselines: U-Net-6L, SfSNet~\cite{sengupta2018sfsnet}, InverseFaceNet~\cite{kim2018inversefacenet}, and GAN2X~\cite{pan2022gan2x}. While baselines struggle to disentangle cast shadows and surface pigmentation, our Stage I (MAGINet) produces more even, photometrically consistent albedo maps. RefinementNet further enhances fine structures such as wrinkles, moles, and eyebrows, absent in earlier stages.  Additionally, Figure~\ref{fig:relight} illustrates the practical advantage of our proposed method by showing realistic relighting results under different illumination conditions, specifically daylight and night lighting scenarios. Our method maintains accurate facial structures and consistent shading across varying lighting conditions, highlighting its robustness and potential for practical applications in realistic rendering tasks.

\begin{figure}[!htbp]
\centering
\includegraphics[scale=0.12]{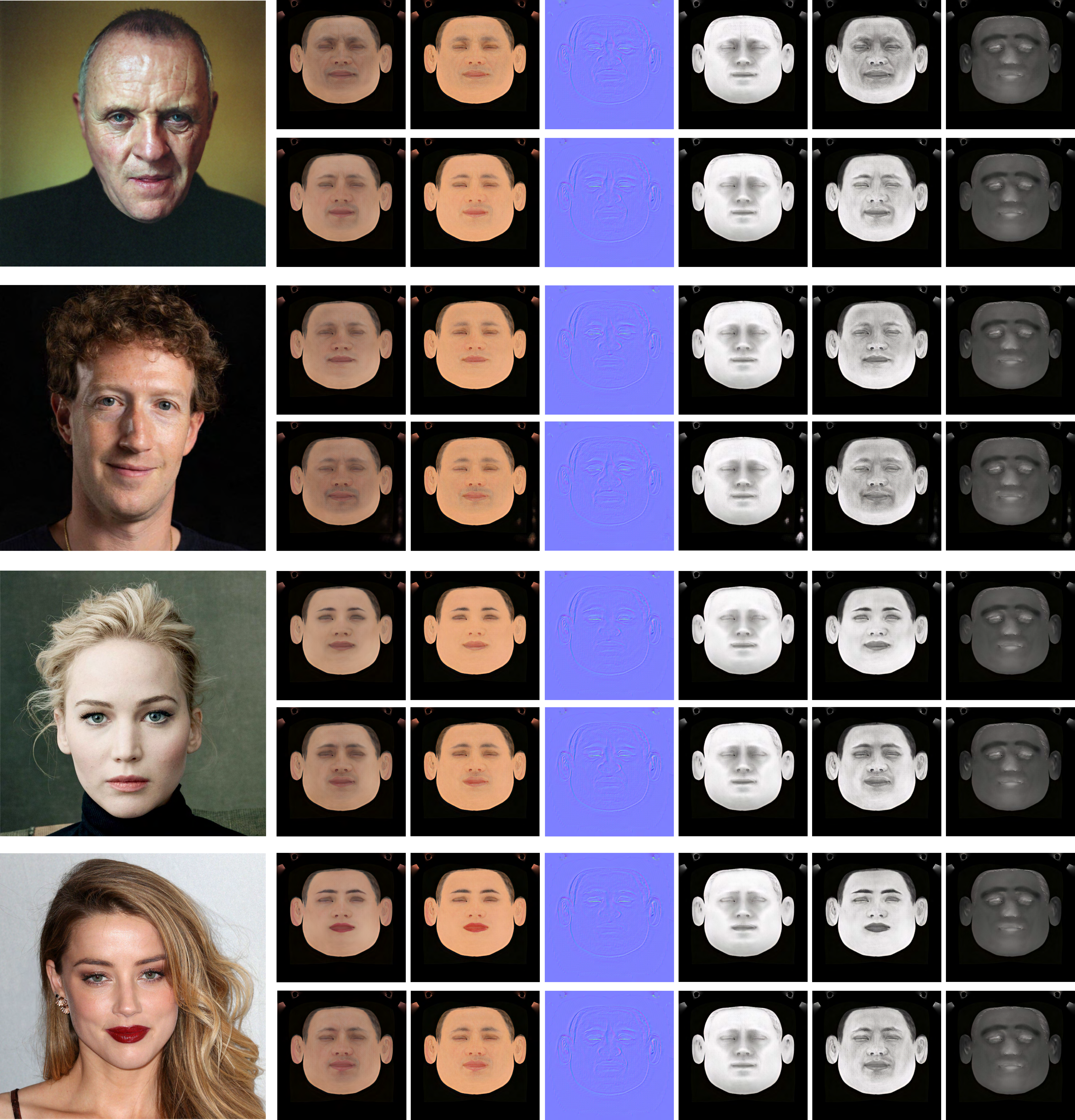}
\caption{Intrinsic decomposition outputs from Stage III (Pix2PixHD), showing (left → right) predicted light-normalised albedo, raw diffuse colour (with residual lighting), normal, ambient occlusion, specular reflection and translucency. Our method produces visually coherent and physically plausible results with significantly more detailed occlusion and sharper normal transitions around facial contours compared to the RGB2AO~\cite{inoue2020rgb2ao} baseline (bottom row).}
\label{fig:qual_passes}
\end{figure}

\paragraph{Rendering pass synthesis.}
Figure~\ref{fig:qual_passes} illustrates the full set of predicted intrinsic passes generated by Stage III, including ambient occlusion, normals, specular, translucency, and raw diffuse colour (with residual lighting). Our pipeline produces visually coherent and physically plausible outputs. Compared to RGB2AO~\cite{inoue2020rgb2ao} baselines (bottom row), our method yields significantly more detailed occlusion and sharper normal transitions around facial contours.

\section{Conclusion and Discussion}\label{sec:conclusion}

In this study, we introduced a novel multi-stage neural rendering pipeline combining MAGINet, RefinementNet, and Pix2PixHD~\cite{wang2018Pix2PixHD} to achieve intrinsic image decomposition of facial portraits. The pipeline substantially improves the estimation of albedo, ambient occlusion, surface normals, specular highlights, translucency, and diffuse reflection on the challenging FFHQ--UV–Intrinsics dataset~\cite{dib2024mosar}.

Our methodology leverages progressive refinement through its multi-stage architecture, enabling high-quality decomposition by iteratively correcting and enhancing previous outputs. In particular, RefinementNet mitigates initial reconstruction artefacts from MAGINet and yields sharper, more coherent intrinsic layers.

However, several limitations remain. Facial hair introduces complex geometry and reflectance that often degrade normal and specular estimation. Similarly, occlusions such as bangs covering the forehead or eyes disrupt accurate albedo and ambient occlusion recovery. Performance also drops for darker skin tones, reflecting dataset imbalance and limited diversity. Furthermore, as the ``ground-truth'' maps in FFHQ-UV-Intrinsics~\cite{wang2018Pix2PixHD} are themselves produced by the MoSAR inverse-rendering pipeline, any modelling bias or error in MoSAR propagates into our supervision; thus, the recovered passes should be viewed as data-consistent rather than absolute physical measurements.

Future work will focus on improving robustness to complex occlusions, handling facial hair and diverse skin tones, and validating against light-stage scans to tighten the link between data-driven intrinsics and physical reality. Exploring training strategies or augmentations that capture a wider range of lighting conditions and facial variations should further enhance real-world reliability.

\section*{Statements and Declarations}

\subsection*{Competing Interests}
The author has no relevant financial or non-financial interests to disclose.

\subsection*{Funding}
This research received no external funding. 

\subsection*{Author Contributions}
The sole author performed all work related to the conception, design, implementation, analysis, and writing of this study.

\subsection*{Data Availability}
The code and trained models are available from the corresponding author upon request.

\bibliography{sn-bibliography}

\end{document}